# Me-LLaMA: Medical Foundation Large Language Models for Comprehensive Text Analysis and Beyond

**Authors:** Qianqian Xie, PhD[1,*], Qingyu Chen, PhD[1,*], Aokun Chen, PhD[2,*], Cheng Peng, PhD[2], Yan Hu, MS[3], Fongci Lin, PhD[1], Xueqing Peng, PhD[1], Jimin Huang, MS[1], Jeffrey Zhang, PhD[1], Vipina Keloth, PhD[1], Xinyu Zhou, BS[1], Lingfei Qian, PhD[1], Huan He, PhD[1], Dennis Shung, MD, PhD[1,4], Lucila Ohno-Machado, MD, PhD[1], Yonghui Wu, PhD[2], Hua Xu, PhD[1,+], and Jiang Bian, PhD[2,+]

**Affiliations:**

[1]Department of Biomedical Informatics and Data Science, Yale School of Medicine, Yale University, New Haven, CT, USA

[2]Department of Health Outcomes and Biomedical Informatics, College of Medicine, University of Florida, Gainesville, FL, USA

[3]School of Biomedical Informatics, University of Texas Health Science, Center at Houston, Houston, TX, USA

[4]Department of Medicine (Digestive Diseases), Yale School of Medicine, Yale University, New Haven, CT, USA

[*]These authors contributed equally to this work.

[+]Corresponding authors: hua.xu@yale.edu, bianjiang@ufl.edu

Address:

100 College Street, New Haven, Connecticut 06510

1889 Museum Rd, Gainesville, Florida 32610

## Abstract

**Background** Recent advancements in large language models (LLMs) like ChatGPT and LLaMA have shown promise in medical applications, though their performance in medical language understanding still requires enhancement. This study aims to develop foundational medical LLMs by training open-source LLaMA models with large-scale, domain-specific datasets to enhance their efficacy across a variety of medical text analysis tasks and medical diagnosis.

**Methods** We developed Me-LLaMA, a new medical LLM family that includes foundation models – Me-LLaMA 13/70B, and their chat-enhanced versions, through continual pre-training and instruction tuning of LLaMA2 using both biomedical literature and clinical notes. Me-LLaMA utilized the largest and most comprehensive medical data, including 129B pre-training tokens and 214K instruction tuning samples from diverse biomedical and clinical data sources. Training the 70B models required substantial computational resources, exceeding 100,000 A100 GPU hours. We applied Me-LLaMA to six medical text analysis tasks and evaluated its performance on 12 benchmark datasets. To further assess Me-LLaMA's potential clinical utility, we evaluated its performance on complex clinical case diagnosis compared with other commercial LLMs, using both automatic and human evaluations.

**Results** Me-LLaMA models outperform LLaMA, and other existing open-source medical LLMs in both zero-shot and supervised learning settings for most text analysis tasks. With task-specific instruction tuning, Me-LLaMA models also surpass leading commercial LLMs, outperforming ChatGPT on 7 out of 8 datasets and GPT-4 on 5 out of 8 datasets. Moreover, for diagnosing complex clinical cases, Me-LLaMA's performance is comparable to ChatGPT and GPT-4.

**Conclusion** Domain-specific data is crucial for building medical foundation LLMs that enhance diverse downstream text analysis tasks and medical applications. The substantial computing costs associated with training such models require careful consideration of training strategies (pre-training vs. fine-tuning). Me-LLaMA models are now publicly available through appropriate user agreements, making them a valuable resource for advancing medical AI applications.

## INTRODUCTION

Large language models (LLMs) have shown great potential in improving medical applications such as clinical documentation, diagnostic accuracy, and patient care management.[1,2,3] Commercial LLMs, such as ChatGPT [4] and GPT-4,[5] are usually closed-source, which limits flexible customization and easy accessibility that are often required for medical applications. Meanwhile, recent research efforts have pivoted towards the development of open-source LLMs, with LLaMA models[6,7] pioneering in the general domain. Nevertheless, general domain LLMs may not always have specialized medical knowledge that is essential for accurate and reliable applications in medicine, probably because they are trained mainly on datasets from the general domain.[8]

In response, there has been much attention on developing open-source medical foundation LLMs using medical data.[2,3,9] For example, GatotronGPT[9] is a medical LLMs pre-trained from scratch using the GPT-3 architecture, with model sizes of 5B and 20B parameters. As novel open-source general domain LLMs with larger model sizes and superior performance emerged, such as the LLaMA models, researchers began developing medical LLMs by adapting these open-domain models and supplying them with more medical data. This is typically achieved through two approaches: continual pre-training or fine tuning with instructions. Continual pre-training enhances the medical knowledge of open-domain LLMs with extensive medical data, while instruction tuning improves their instruction-following and generalization abilities. Notable medical LLMs developed through these methods include Meditron[10] and Clinical LLaMA,[11] which continue pretraining LLaMA2 and LLaMA models with extensive biomedical literature or clinical notes respectively. Additionally, models like MedAlpaca,[3] ChatDoctor,[12] and AlpaCare[13] were developed by instruction fine-tuning LLaMA2 and LLaMA models. PMC-LLaMA[2] stands out as the only model using both continual pretraining and instruction tuning, based on LLaMA models.

Despite advancements made by existing models, significant challenges remain for developing open-source medical foundation LLMs: (1) Few models (except PMC-LLaMA) employed both continual pretraining and instruction fine-tuning, probably due to the expensive computational costs associated with model training (especially for pre-training). (2) Only one model (Clinical LLaMA) used clinical notes from electronic health records, which is crucial for real-world clinical applications as it provides context-specific information from direct patient care. Furthermore, none of the existing models used both biomedical literature and clinical notes, which is one of the goals of this project. By combining biomedical literature and clinical notes, we generated the largest biomedical pre-training dataset (129B tokens), compared to the previous efforts (i.e., 79B tokens in PMC-LLaMA as the highest, see Table 1). (3) Most models have focused predominantly on evaluating QA tasks, making it difficult to assess the generalizability of those foundation models on other medical text analysis tasks such as information extraction and text classification.

**Table 1**. The comparison of Me-LLaMA models and existing open source medical LLMs.

| Model | Backbone | Model size | Biomedical literature | Clinical notes | Continual pre-training (# of tokens) | Instruction tuning (# of instructions) | Evaluation tasks | Release date |
|---|---|---|---|---|---|---|---|---|
| MedAlpaca | LLaMA | 7/13B | ✓ | ✗ | - | 160K | QA | 04/14/2023 |
| ChatDoctor | LLaMA2 | 7B | ✓ | ✗ | - | 100K | QA | 05/24/2023 |
| AlpaCare | LLaMA | 7/13B | ✓ | ✗ | - | 52K | QA, Summarization | 10/23/2023 |
| Clinical LLaMA | LLaMA | 7B | ✗ | ✓ | - | - | Classification | 07/06/2023 |
| Meditron | LLaMA2 | 7/70B | ✓ | ✗ | 48B | - | QA | 11/27/2023 |
| PMC-LLaMA | LLaMA | 7/13B | ✓ | ✗ | 79B | 514K | QA | 04/27/2023 |
| **Me-LLaMA** | LLaMA2 | 13/70B | ✓ | ✓ | 129B | 214K | QA, NER, RE, Classification, Summarization, NLI, Medical Diagnosis | 06/05/2024 |

To address the above challenges, we propose Me-LLaMA, a new family of open-source medical LLMs, including foundation models (Me-LLaMA 13/70B), and their chat-enhanced versions (Me-LLaMA 13/70B-chat). Developed through extensive continual pre-training and instruction tuning of LLaMA2 models, Me-LLaMA utilizes the largest and most comprehensive medical data to date, combining 129B pre-training tokens and 214K instruction tuning samples, from scientific literature, clinical guidelines, and clinical notes of EHRs. This extensive dataset enables Me-LLaMA to excel in a wide range of medical text analysis tasks and real world clinical tasks. Compared to existing studies, we perform the most comprehensive evaluation, covering six critical text analysis tasks: question answering, relation extraction, named entity recognition, text classification, text summarization, and natural language inference, using 12 evaluation datasets from both biomedical and clinical domains. Our extensive evaluation shows that Me-LLaMA models achieve overall superior performance compared to existing open-source medical LLMs in both zero-shot and supervised learning settings. With task-specific instruction tuning, Me-LLaMA models outperform leading commercial LLMs including ChatGPT on 7 out of 8 datasets and GPT-4 on 5 out of 8 datasets.

To further assess Me-LLaMA's potential clinical utility, we also evaluated Me-LLaMA models on the complex clinical case diagnosis task and compared its performance with other commercial LLMs, using both automatic and human evaluation. Our evaluation shows that Me-LLaMA's performance is comparable to ChatGPT and GPT-4, despite their significantly larger model sizes. Me-LLaMA models have been approved for releasing to the public on PhysioNet through appropriate DUAs, and our datasets and evaluation scripts have been released publicly on GitHub. We hope these resources will benefit the medical AI community, fostering further innovation and development in this critical field.

**METHODS**

We utilized LLaMA2 as the backbone model and developed Me-LLaMA through the process of continual pre-training and instruction tuning of LLaMA2, using 129B tokens and 214K instruction tuning samples from general, biomedical, and clinical domains. Figure 1 shows an overview of our study.

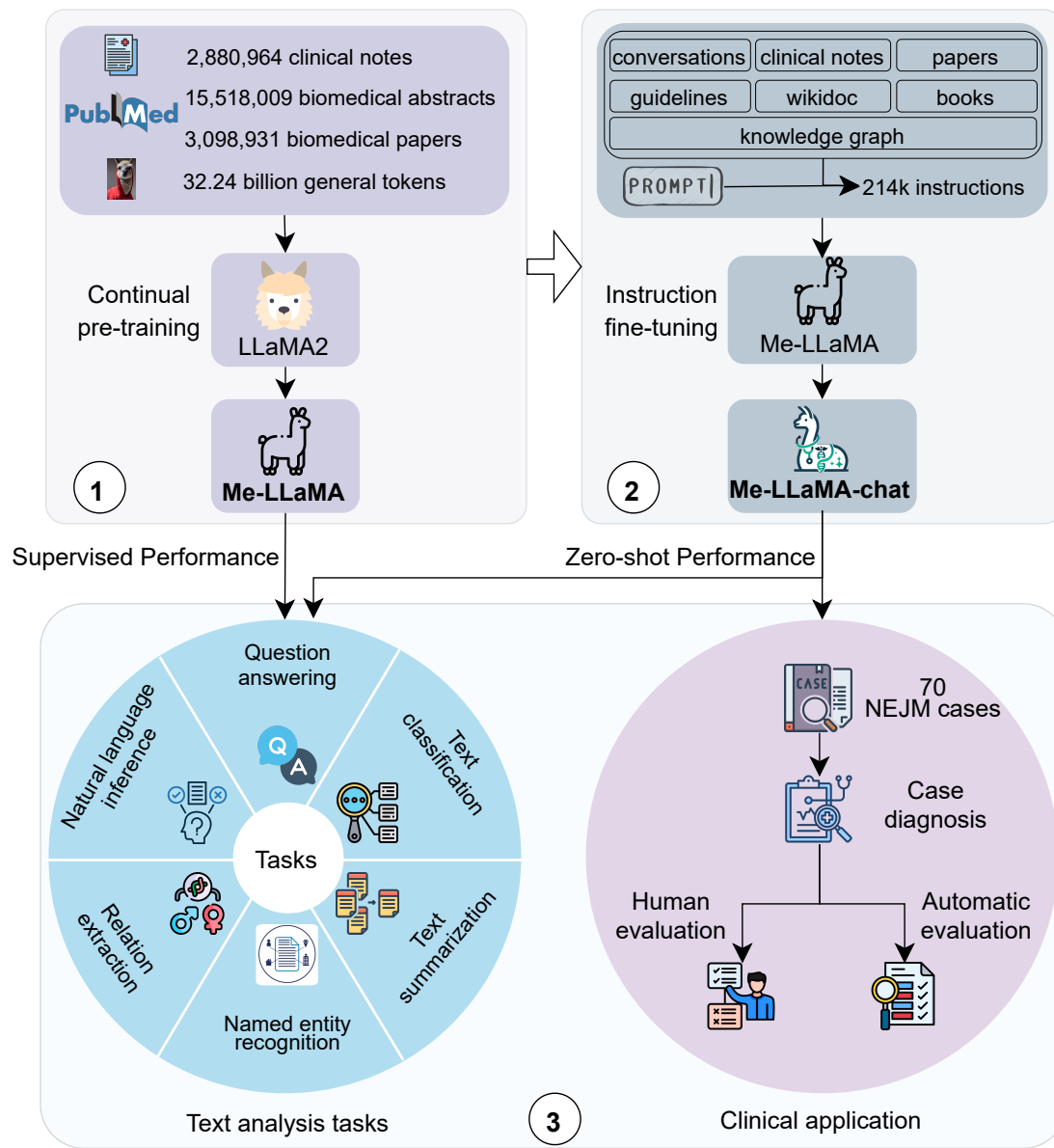

**Figure 1**. Overview of the study. Our study has three main components including pre-training, instruction fine-tuning and evaluation. Pre-training: we firstly developed the Me-LLaMA base models by continual pre-training LLaMA2 with 129 billion tokens from mixed pre-training text data. Instruction fine-tuning: Me-LLaMA-chat models were further developed by instruction tuning Me-LLaMA base models with 214K instructions. Evaluation: Finally, we evaluated the Me-LLaMA base models in a supervised learning setting across six text analysis tasks, and the Me-LLaMA-chat models in a zero-shot setting on both text analysis tasks and a clinical diagnosis task.

**Continual Pre-Training Data**

To effectively adapt backbone LLaMA2 models for the medical domain through continual pre-training, we developed a mixed continual pre-training dataset, comprised of biomedical literature, clinical notes, and general domain data. It integrates over 3 million full biomedical articles from PubMed Central and over 15 million paper abstracts from PubMed, sourced from the Pile dataset.[14] To incorporate real-world clinical scenarios and reasoning, we included de-identified free-text clinical notes from MIMIC-III,[15] MIMIC-IV,[16] and MIMIC-CXR.[17] Moreover, to avoid the model forgetting acquired general knowledge, we incorporated a subset from the RedPajama[18] dataset, a replication of LLaMA2's pre-training data. The dataset was structured with a 15:1:4 ratio of biomedical, clinical, to general domain data and contains a total of 129 billion tokens, making it the largest pre-training dataset in the medical domain currently available.

**Medical Instruction Tuning Data**

To enhance our model's ability to follow instructions and generalize across diverse medical tasks, we further developed a novel medical instruction tuning dataset with 214,595 high-quality samples from a wide array of data sources. This dataset stands out from those used in existing medical LLMs due to its comprehensive coverage of both biomedical and clinical domains. Our data sources included biomedical literature, clinical notes, clinical guidelines, wikidoc, knowledge graphs, and general domain data, as shown in Table 2. The diverse tasks aim to refine the model's ability to process and respond to medical information accurately and contextually. Detailed prompts for each data and the data example are shown in Appendix 0.1, Table A.1.

**Table 2**. The overall instruction tuning dataset.

| Task | Type | Source | Size | Copy right |
|---|---|---|---|---|
| General | Conversation | Alpaca[19] | 20,000 | CC-BY-NC 4.0 |
| | | Dolly[20] | | CC-BY-SA-3.0 |
| | | ShareGPT[21] | | Apache-2.0 |
| Biomedical | Conversation | HealthCareMagic[12] | 20,000 | Reserved by HealthCareMagic and Icliniq |
| | | Icliniq[12] | | |
| | Instructions | MedInstruct[13] | 52,000 | CC BY-NC 4.0 |
| | Question Answering | Medical Flash Cards[3] | 34,000 | No commercialized use |
| | | MEDIQA[22] | 2,220 | CC BY 4.0 |
| | | MedicationQA[23] | 690 | CC BY 4.0 |
| | | LiveQA[24] | 634 | CC BY 4.0 |
| | | WikiDocPatient[3] | 5,490 | CC BY-SA 4.0 |
| | | GuidelineQA | 2,000 | Common Crawl (other) |
| | Summarization | PubMed Central | 10,000 | CC BY |
| | Next Sentence Generation | PubMed Central | 20,000 | CC BY |
| | Key words prediction | PubMed Central | 10,000 | CC BY |
| | Causal Relation Detection | PubMed[25] | 2,450 | CC BY |
| | Relation Extraction | UMLS knowledge graph[2] | 10,000 | Openrail |
| Clinical | QA, summarization, classification, mortality prediction | MIMIC-III,[15] MIMIC-IV[16] | 30,000 | PhysioNet credentialed health data use agreement 1.5.0 |

**Training Details**

As shown in Figure 3, we developed the Me-LLaMA 13B and 70B base models by continual pre-training the LLaMA2 13B and 70B models. These base models were then instruction-tuned to create the Me-LLaMA-13B-chat and Me-LLaMA-70B-chat models.

**Me-LLaMA base models - continual pretraining LLaMA2**: This phase aims to adapt LLaMA2 models to better understand and generate text relevant to the medical context using the pre-training datasets we constructed. The training involves sequences of medical texts, where the model learned to predict the next token $x_{i+1}$ in a sequence $\{x_1, x_2, \cdots, x_n\}$, maximizing the likelihood $\mathcal{L}(\Theta) = \sum_{i=1}^{n-1} \log P_\Theta(x_{i+1}|x_1, x_2, \cdots, x_i)$, where $\Theta$ is the parameter set of LLaMA2 models. This training was executed on the University of Florida's HiPerGator AI supercomputer with 160 A100 80GB GPUs. We employed the AdamW optimizer with hyperparameters set to $\beta_1$ to 0.9 and $\beta_2$ to 0.95, alongside a weight decay of 0.00001 and a learning rate of 8e-6. We used a cosine learning rate scheduler with a 0.05 warmup ratio for gradual adaptation to training complexity and bf16 precision for computational efficiency. Gradient accumulation was set to 16 steps, and training was limited to one epoch. We utilized DeepSpeed[26] for model parallelism.

**Me-LLaMA chat models - instruction fine-tuning Me-LLaMA**: We further fine-tuned Me-LLaMA base models, using the developed 214k instruction samples. The training objective is to maximize the likelihood: $\mathcal{L}(\Theta) = argmax \sum_{(x^i, y^i) \in (X,Y)} \log p(y^i|x^i; \Theta)$, where $x^i$ represents the input instruction, $y^i$ is the ground truth response, and $\Theta$ is the parameter set of Me-LLaMA. Executed using 8 A100 GPUs, the fine-tuning process was set to run for 3 epochs with a learning rate of 1e-5. We used a weight decay of 0.00001 and a warmup ratio of 0.01 for regularization and gradual learning rate increase. We utilized LoRA-based[27] parameter-efficient fine-tuning.

**Evaluation Benchmark**

**Biomedical and clinical NLP tasks**: Existing studies[2,3,10,11] in the medical domain have primarily focused on evaluating the QA task. In this study, we build an extensive medical evaluation benchmark (MIBE), encompassing six critical text analysis tasks: QA, NER, RE, Text Classification, Text Summarization and NLI. These tasks collectively involve 12 datasets meticulously sourced from biomedical, and clinical domains as shown in Table 3.

**Table 3**. Details of data splits and evaluation metrics of each dataset in the evaluation benchmark.

| Data | Task | Train | Valid | Test | Evaluation |
|---|---|---|---|---|---|
| PubMedQA*[28] | QA | 190,143 | 21,126 | 500 | Accuracy, Macro-F1 |
| MedQA[29] | QA | 10,178 | 1,272 | 1,273 | Accuracy, Macro-F1 |
| MedMCQA*[30] | QA | 164,540 | 18,282 | 4,183 | Accuracy, Macro-F1 |
| EmrQA[31] | QA | 122,326 | 30,581 | 26,804 | Exact match, F1 |
| i2b2[32] | NER | 6,0875 | 7,400 | 7,451 | Entity-level Macro-F1 |
| DDI[33] | RE | 18,779 | 7,244 | 5,761 | Macro-F1 |
| HoC[34] | Classification | 1,108 | 157 | 315 | Label-wise Macro-F1 |
| MTSample[35] | Classification | 4,999 | 500 | 999 | Accuracy, Macro-F1 |
| PubMed[36] | Summarization | 117,108 | 6,631 | 6,658 | Rouge, BERTScore |
| MIMIC-CXR[17] | Summarization | 122,014 | 957 | 1,606 | Rouge, BERTScore |
| BioNLI[37] | NLI | 5,544 | 5,000 | 6,308 | Accuracy, Macro-F1 |
| MedNLI[38] | NLI | 11,232 | 1,422 | 1,395 | Accuracy, Macro-F1 |

**Complex clinical case diagnosis task:** we further assessed the effectiveness of Me-LLaMA in diagnosing complex clinical cases, a critical task given the increasing burden of diseases and the need for timely and accurate diagnosis to support clinicians. Recent studies demonstrate that LLMs have the potential to address this challenge.[39] Specifically, we evaluated the diagnostic accuracy of Me-LLaMA on 70 challenging medical cases from the New England Journal of Medicine clinicopathologic conferences (NEJM CPCs) published between January 2021 and December 2022, as collected from an existing study.[39] The NEJM CPCs are well-known for their unique and intricate clinical cases, which have long been used as benchmarks for evaluating challenging medical scenarios. In line with previous research,[39, 40] we employed automatic evaluations based on top-K (where k=1,2,3,4,5) accuracy, defined as the percentage of cases where the correct diagnosis appeared within the top-K positions of the differential diagnosis list predicted by the assessed models. We utilized GPT-4o, a state-of-the-art (SOTA) LLM, to automatically assess whether each diagnosis from the model's differential diagnosis list matched the gold standard final diagnosis, consistent with these prior studies. Existing studies[40] have shown that LLM-based automatic calculation of top-K accuracy is comparable to human evaluation. Besides automatic evaluation, we had a clinician specializing in internal medicine perform a manual evaluation of top-k accuracy (k=1, 5). For more details on data processing, automatic evaluation, and human evaluation, see Appendix A.3.

**Evaluation Settings**

We evaluated Me-LLaMA at two evaluation settings including zero-shot and supervised learning to evaluate their performance and generalization ability across various tasks compared to baseline models.

**Supervised Learning**

In the supervised learning setting, we evaluated Me-LLaMA 13/70B base models' performances adapted to downstream tasks. We conducted the task-specific finetuning on Me-LLaMA base models (Me-LLaMA task-specific) with each training set of assessed datasets in Table 6, and then assessed the performance of Me-LLaMA task-specific models on test datasets. We employed the AdamW optimizer. For datasets with fewer than 10,000 training samples, we fine-tuned the models for 5 epochs, while for larger datasets, the fine-tuning was conducted for 3 epochs. A uniform learning rate of 1e-5 was used across all datasets. Our baseline models including LLaMA2 Models (7B/13B/70B)[7]: they are open-sourced LLMs released by Meta AI. PMC-LLaMA 13B[2] is a biomedical LLM continually pre-trained on biomedical papers and medical books. Meditron7B/70B[10]: they are medical LLMs based on LLaMA2-7B/70B, continual pre-trained with a mix of clinical guidelines, medical papers and abstracts.

**Zero-shot Learning**

We assessed our Me-LLaMA 13/70B-chat models' zero-shot learning capabilities, which are key for new task understanding and response without specific prior training. We compared our models and baseline models' zero-shot, using standardized prompts (detailed in Table A.2 shown in Appendix 0.2) for each test dataset from Table 2. We compared Me-LLaMA 13/70B-chat models with the following baseline models: ChatGPT/GPT-4[4,5]: SOTA commercialized LLMs. We used the version of "gpt-3.5-turbo-0301" for ChatGPT, and the version of "gpt-4-0314" for GPT-4. LLaMA2-7B/13B/70B-chat[7] models were adaptations of the LLaMA2 series, optimized for dialogue and conversational scenarios. Medalpaca-7B/13B[3] models were based on LLaMA-7B/13B, specifically fine-tuned for tasks in the medical domain. The PMC-LLaMA-13B-chat[2] model is an instruction-tuned medical LLM based on PMC-LLaMA-13B. The AlpaCare-13B[13] model is specifically tailored for clinical tasks based on LLaMA-2 13B by instruction tuning. Meditron 70B[10] is a medical LLM, continually pre-trained with a mix of clinical guidelines, biomedical papers, and abstracts based on LLaMA2 70B.

**RESULTS**

**Overall Performance: Medical Text Analysis**

Table 4 compares the performance of our Me-LLaMA 13/70B foundation models against other open LLMs in the supervised setting. We can observe that the Me-LLaMA 13B model surpassed the similar-sized medical foundation model PMC-LLaMA 13B on 11 out of 12 datasets and outperformed the general foundation model LLaMA2 13B on 10 out of 12 datasets. Moreover, it is noticed that the Me-LLaMA 13B model was competitive with LLaMA2 70B and Meditron 70B, which have significantly larger parameter sizes, on 8 out of 12 datasets. As for 70B models, Me-LLaMA 70B achieved the best performance on 9 out of 12 datasets, when benchmarked against LLaMA2 70B and Meditron 70B.

**Table 4**. The supervised fine-tuning performance of various open source LLMs on six tasks.

| Task | Dataset | Metric | LLaMA2 13B | PMC-LLaMA 13B | Me-LLaMA 13B | LLaMA2 70B | Meditron 70B | Me-LLaMA 70B |
|---|---|---|---|---|---|---|---|---|
| Question answering | PubMedQA | Acc | 0.800 | 0.778 | **0.802** | 0.800 | 0.800* | **0.814** |
| | | Macro-F1 | 0.560 | 0.544 | **0.562** | 0.560 | - | **0.572** |
| | MedQA | Acc | 0.467 | 0.456 | **0.493** | 0.598 | 0.607* | **0.623** |
| | | Macro-F1 | 0.465 | 0.454 | **0.487** | 0.595 | - | **0.621** |
| | MedMCQA | Acc | 0.527 | 0.548 | **0.557** | 0.626 | **0.651*** | 0.643 |
| | | Macro-F1 | 0.524 | 0.545 | **0.551** | 0.625 | - | **0.640** |
| | EmrQA | Acc | 0.789 | 0.810 | **0.857** | 0.847 | 0.850 | **0.854** |
| | | F1 | 0.730 | 0.738 | **0.751** | 0.751 | 0.751 | **0.751** |
| Named entity recognition | i2b2 | Macro-F1 | 0.904 | 0.901 | **0.906** | **0.913** | 0.908 | 0.910 |
| Relation extraction | DDI | Macro-F1 | **0.622** | 0.622 | 0.559 | 0.746 | 0.737 | **0.779** |
| Classification | HoC | Macro-F1 | **0.696** | 0.422 | 0.684 | 0.818 | 0.702 | **0.841** |
| | MTsample | Macro-F1 | 0.430 | 0.345 | **0.451** | 0.458 | 0.284 | **0.544** |
| Summarization | PubMed | R-L | 0.191 | 0.091 | **0.197** | **0.211** | 0.197 | 0.209 |
| | | BERTS | 0.663 | 0.516 | **0.679** | 0.689 | 0.677 | **0.700** |
| | MIMIC-CXR | R-L | 0.437 | 0.139 | **0.453** | 0.440 | 0.458 | **0.476** |
| | | BERTS | 0.816 | 0.694 | **0.821** | 0.813 | 0.824 | **0.828** |
| Natural language inference | BioNLI | Macro-F1 | 0.409 | 0.332 | **0.447** | 0.447 | 0.444 | **0.566** |
| | MedNLI | Macro-F1 | 0.881 | 0.868 | **0.903** | 0.884 | 0.897 | **0.916** |

*The performance of Meditron 70B on the PubMedQA, MedQA, and MedMCQA datasets is cited from the meditron paper[10] to have a fair comparison.

Table 5 shows the zero-shot performance of Me-LLaMA chat models and other instruction tuned open LLMs with chat ability on various tasks. Among 13B models, Me-LLaMA 13B-chat outperformed LLaMA2 13B-chat, PMC-LLaMA-chat, Medalpaca 13B in almost all 12 datasets. Me-LLaMA outperformed AlpaCare-13B in 9 out of 12 datasets. Among models with 70B parameters, Me-LLaMA 70B-chat consistently outperformed LLaMA2-70B-chat on 11 out of 12 datasets. It is worth noting that Me-LLaMA13B-chat showed better performance than LLaMA2-70B-chat—a model with a significantly larger parameter size—on 6 out of 12 datasets and was competitive with the LLaMA2-70B-chat in 3 out of 6 remaining datasets.

**Table 5**. The zero-shot performance of various open source LLMs with chat capability.

| Task | Dataset | Metric | LLaMA2-13B-chat | PMC-LLaMA-chat | Medalpaca-13B | AlpaCare-13B | Me-LLaMA 13B-chat | LLaMA2-70B-chat | Me-LLaMA 70B-chat |
|---|---|---|---|---|---|---|---|---|---|
| Question answering | PubMedQA | Accuracy | 0.546 | 0.504 | 0.238 | 0.538 | **0.700** | 0.668 | **0.768** |
| | | Macro-F1 | 0.457 | 0.305 | 0.192 | 0.373 | **0.504** | 0.477 | **0.557** |
| | MedQA | Accuracy | 0.097 | 0.207 | 0.143 | 0.304 | **0.427** | 0.376 | **0.523** |
| | | Macro-F1 | 0.148 | 0.158 | 0.102 | 0.281 | **0.422** | 0.367 | **0.521** |
| | MedMCQA | Accuracy | 0.321 | 0.212 | 0.205 | 0.385 | **0.449** | 0.339 | **0.539** |
| | | Macro-F1 | 0.243 | 0.216 | 0.164 | 0.358 | **0.440** | 0.273 | **0.538** |
| | EmrQA | Accuracy | 0.001 | **0.053** | 0.000 | 0.001 | 0.048 | 0.050 | **0.119** |
| | | F1 | 0.098 | 0.304 | 0.040 | 0.198 | **0.307** | 0.251 | **0.346** |
| Named entity recognition | i2b2 | Macro-F1 | 0.143 | 0.091 | 0.000 | **0.173** | 0.166 | 0.321 | **0.329** |
| Relation extraction | DDI | Macro-F1 | 0.090 | 0.147 | 0.058 | 0.110 | **0.214** | 0.087 | **0.283** |
| Classification | HoC | Macro-F1 | 0.228 | 0.184 | 0.246 | 0.267 | **0.335** | 0.309 | **0.544** |
| | MTsample | Macro-F1 | 0.133 | 0.083 | 0.003 | **0.273** | 0.229 | 0.254 | **0.384** |
| Summarization | PubMed | Rouge-L | 0.161 | 0.028 | 0.014 | **0.167** | 0.116 | **0.192** | 0.169 |
| | | BERTS* | 0.671 | 0.128 | 0.117 | **0.671** | 0.445 | **0.684** | 0.678 |
| | MIMIC-CXR | Rouge-L | 0.144 | 0.139 | 0.010 | 0.134 | **0.400** | 0.131 | **0.418** |
| | | BERTS* | 0.704 | 0.694 | 0.502 | 0.702 | **0.797** | 0.696 | **0.787** |
| Natural language inference | BioNLI | Macro-F1 | 0.173 | 0.159 | 0.164 | 0.170 | **0.195** | 0.297 | **0.436** |
| | MedNLI | Macro-F1 | 0.412 | 0.175 | 0.175 | 0.275 | **0.472** | 0.515 | **0.675** |

*BERTS: BERTScore.[41]

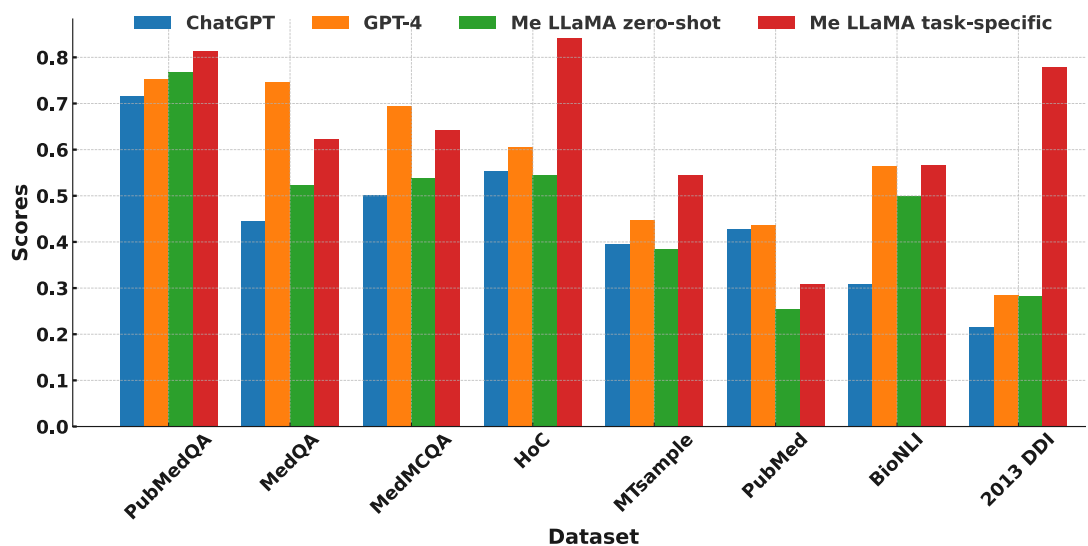

**Figure 2**. The performance comparison of Me-LLaMA models in both zero-shot (Me-LLaMA zero-shot) and supervised learning (Me-LLaMA task-specific) settings, against the zero-shot performance of ChatGPT and GPT-4.

Figure 2 further compares the performance of Me-LLaMA models in the zero-shot and supervised learning setting, against ChatGPT and GPT-4. Due to privacy concerns, which preclude the transmission of clinical datasets with patient information to ChatGPT and GPT-4, we conducted our comparison across 8 datasets that are not subject to these limitations. The results of ChatGPT and GPT-4 on three QA datasets are referenced from the OpenAI's paper.[1] We compared the Rouge-1[42] score for the summarization dataset PubMed, the accuracy score for three QA datasets, and the Macro-F1 score for the remaining datasets. With task-specific supervised fine-tuning, Me-LLaMA models surpassed ChatGPT on 7 out of 8 datasets and excelled GPT-4 on 5 out of 8 datasets. In the zero-shot setting, Me-LLaMA models outperformed ChatGPT on 5 datasets; but it fell short on 7 datasets, when compared with GPT-4. It's crucial to highlight that Me-LLaMA's model size is significantly smaller—13/70B parameters versus at least 175B for ChatGPT and GPT-4. Despite this size discrepancy, Me-LLaMA models have showcased an impressive performance and a strong ability for supervised learning and zero-shot learning across a broad spectrum of medical tasks, underscoring its efficiency and potential in the field.

**Clinical Application: Complex Clinical Case Diagnosis**

Figure 3 shows the top-K (1≤K≤5) accuracy of Me-LLaMA-70B-chat, ChatGPT, GPT-4, and LLaMA2-70B-chat, in the complex clinical case diagnosis task. We can see Me-LLaMA-70B-chat model achieved comparable performance with GPT-4 and ChatGPT, and significantly outperforms LLaMA2-70B-chat. The human evaluation result in Figure 4 again shows that Me-LLaMA-70B-chat outperformed GPT-4 in both top-1 and top-5 accuracy. These results demonstrated the potential of Me-LLaMA models for challenging clinical applications.

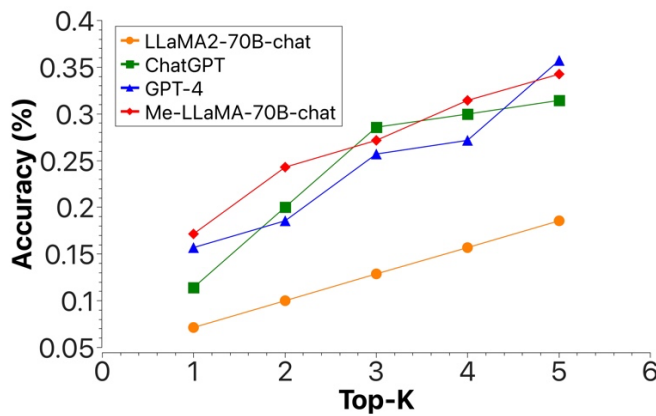

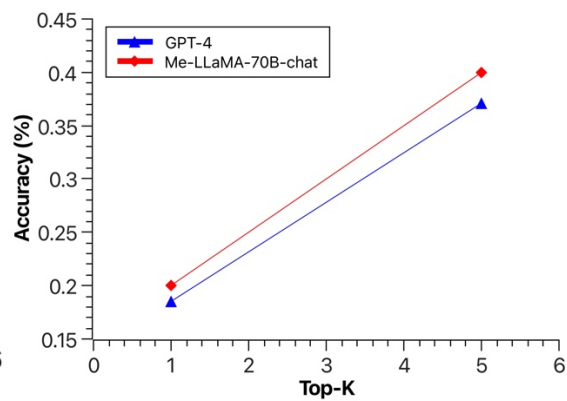

**Figure 3**. The top-k (1<=k<=5) accuracy of different LLMs in complex clinical case diagnosis, with automatic evaluation.

**Figure 4**. The top-1 and top-5 accuracy of Me-LLaMA-70B-chat and GPT-4 in complex clinical case diagnosis, with human evaluation.

**Ablation Study: Impact of Continual Pretraining and Instruction Tuning**

Table 6 compares the zero-shot performances of Me-LLaMA models and their backbone models LLaMA2, to illustrate the impact of continual pre-training and instruction tuning. Table 3 clearly demonstrates that both continual pre-training and instruction tuning significantly enhanced the zero-shot capabilities of models. For example, the Me-LLaMA 70B model showed an improvement in performance ranging from 2. 1% to 55% across various datasets in comparison to the LLaMA2 13B model, highlighting the benefits of continual pre-training. The instruction tuning was also found to provide great increases in zero-shot performance. For instance, the Me-LLaMA-70B-chat model displayed enhancements in performance from 3.7% to 41.9% relative to the Me-LLaMA 70B foundation model, which had not undergone instruction tuning. This enhancement suggests the critical role of instruction finetuning in boosting the model's ability to leverage context in learning tasks, even without supervised fine-tuning and prior examples.

**Table 6**. The comparison of zero-shot performances among Me-LLaMA models and their backbone models LLaMA2.

| Dataset | Metric | LLaMA2 13B (backbone) | Me-LLaMA 13B (backbone + pre-train) | Me-LLaMA-13B-chat (backbone + pre-train + instruction tuning) | LLaMA2 70B (backbone) | Me-LLaMA 70B (backbone + pre-train) | Me-LLaMA-70B-chat (backbone + pre-train + instruction tuning) |
|---|---|---|---|---|---|---|---|
| PubMedQA | Acc | 0.216 | 0.266 | **0.700** | 0.132 | 0.682 | **0.768** |
| | Macro-F1 | 0.177 | 0.250 | **0.504** | 0.152 | 0.520 | **0.557** |
| MedQA | Acc | 0.000 | 0.000 | **0.427** | 0.005 | 0.281 | **0.523** |
| | Macro-F1 | 0.000 | 0.000 | **0.422** | 0.009 | 0.350 | **0.521** |
| MedMCQA | Acc | 0.003 | 0.003 | **0.449** | 0.012 | 0.447 | **0.539** |
| | Macro-F1 | 0.006 | 0.005 | **0.440** | 0.024 | 0.396 | **0.538** |
| EmrQA | Acc | 0.000 | 0.005 | **0.048** | 0.000 | 0.021 | **0.119** |
| | F1 | 0.038 | 0.122 | **0.307** | 0.000 | 0.172 | **0.346** |
| i2b2 | Macro-F1 | 0.008 | 0.030 | **0.263** | 0.181 | 0.224 | **0.329** |
| DDI | Macro-F1 | 0.035 | 0.036 | **0.214** | 0.034 | 0.118 | **0.283** |
| HoC | Macro-F1 | 0.253 | 0.210 | **0.335** | 0.255 | 0.252 | **0.544** |
| MTsample | Macro-F1 | 0.042 | 0.072 | **0.229** | 0.066 | 0.226 | **0.384** |
| PubMed | R-L | **0.170** | 0.168 | 0.116 | 0.167 | 0.119 | **0.169** |
| | BERTS | **0.654** | 0.654 | 0.445 | 0.654 | 0.654 | **0.678** |
| MIMIC-CXR | R-L | 0.051 | 0.172 | **0.400** | 0.059 | 0.137 | **0.418** |
| | BERTS | 0.566 | 0.697 | **0.797** | 0.577 | 0.649 | **0.787** |
| BioNLI | Macro-F1 | 0.109 | 0.060 | **0.195** | 0.285 | **0.499** | 0.436 |
| MedNLI | Macro-F1 | 0.172 | 0.206 | **0.472** | 0.265 | 0.256 | **0.675** |

## DISCUSSION

**Model Performance**

We introduced a novel medical LLM family including, Me-LLaMA 13B and Me-LLaMA 70B, which encode comprehensive medical knowledge, along with their chat-optimized variants: Me-LLaMA-13/70B-chat, with strong zero-shot learning ability, for medical applications. These models were developed through the continual pre-training and instruction tuning of LLaMA2 models, using the largest and most comprehensive biomedical and clinical data. Compared to existing studies, we perform the most comprehensive evaluation, covering six critical text analysis tasks. Our evaluations reveal that Me-LLaMA models outperform existing open-source medical LLMs in various learning scenarios, showing less susceptibility to catastrophic forgetting and achieving competitive results against major commercial models including ChatGPT and GPT-4. Our work paves the way for more accurate, reliable, and comprehensive medical LLMs, and underscores the potential of LLMs on medical applications.

In the zero-shot setting, medical LLMs including GPT-4 displayed low performance on certain tasks, e.g., NER and RE, which are also noted by other studies.[43,44] When compared with other NLP tasks with higher performance, we noticed that one of the main reasons for low performance is that LLMs' responses often lacked the conciseness and precision expected, with instances of missing outputs noted. The unexpected outputs also cause significant challenges to automatic evaluation metrics. Therefore, more investigation is needed to further improve medical LLMs' performance across tasks in the zero-shot setting[31] and enhance the automatic assessment of these medical LLMs' zero-shot capabilities. For the complex clinical case diagnosis, the Me-LLaMA-chat model had competitive performance and even outperformed GPT-4 in human evaluation. Existing studies have demonstrated GPT-4 is arguably one of the strongest LLMs in this task.[45] The robust performance of Me-LLaMA showed potential in assisting challenging clinical applications. It is noticed that variations in test sizes and evaluation methods across different studies contribute to the observed differences in performance between GPT-4 in our paper and other studies. We also noted that both the Me-LLaMA-chat model and GPT-4 faced difficulties identifying the correct diagnosis within the top ranks, underscoring the difficulty of this task. Additionally, while the NEJM CPCs offer a rigorous test for these models, they do not encompass the full range of a physician's duties or broader clinical competence. Therefore, complex clinical diagnosis remains a challenging area that demands more effective models and improved evaluation benchmarks to better capture the complexities of real-world clinical scenarios.

**Model Development**

During our model development, we noticed the importance of diversity of the data sources during the pre-training and instruction-tuning phases. Our empirical results revealed that the PMC-LLaMA 13B model, which employed a data mix ratio of 19:1 between medical and general domain data, exhibited around 2.7% performance drop across both general and biomedical tasks. On the other hand, the Meditron models, 7B, and 70B, with a 99:1 mix ratio, demonstrated improvements in biomedical tasks, yet they still saw around 1% declines in the performance of general tasks. In contrast, our models, which adopt a 4:1 ratio, have shown enhancements in their performance for both general and medical tasks. This suggests that the integration of general domain data plays a vital role in mitigating the knowledge-forgetting issue during pre-training.[11,24,25] However, determining the optimal balance between general domain data and specialized medical data is nontrivial, requiring careful empirical analysis. Future studies should examine methods to better determine the optimal ratio.

Our model development also underscores the balance between cost and effectiveness in pre-training versus instruction tuning of LLMs. Pre-training, exemplified by the LLaMA2 70B model, is notably resource-heavy, requiring about 700 hours on 160 A100 GPUs per epoch. Conversely, instruction tuning is far less resource-demanding, needing roughly 70 hours on 8 A100 GPUs per epoch, making it much more affordable than pre-training. Despite this, instruction tuning alone enhanced the performance of the Me-LLaMA-13B-chat model, achieving improvements ranging from 12% to 45% across 11 out of 12 datasets when compared to its backbone model – Me-LLaMA 13B, in the zero-shot setting. This efficiency advocates for prioritizing instruction tuning in scenarios with limited resources, highlighting its potential for cost-effective model enhancement.

**Use of Me-LLaMA Models**

The Me-LLaMA models, available in both 13B and 70B sizes, as well as in base and chat-optimized versions, unlock a wide array of medical applications, guided by the crucial balance between model size and resource availability. The base models serve as robust foundations with extensive medical knowledge, adaptable through supervised fine-tuning for specialized tasks. Conversely, the chat versions excel in instruction-following ability and zero-shot learning, making them highly effective in zero-shot or few-shot learning scenarios. Larger models, like the 70B, provide deeper understanding and more complex reasoning abilities, ideal for comprehensive medical analyses. Yet, their deployment requires significant computing resources, posing challenges in resource-limited settings. On the other hand, the 13B models offer a practical compromise, balancing efficiency with effectiveness, thus ensuring broader accessibility for various applications. Our findings indicate that the Me-LLaMA 13B achieves performance on par with the 70B variant across most datasets, suggesting its viability for diverse medical tasks where computational or financial resources are a concern.

**Limitations**

It is crucial to acknowledge the limitations of the current versions of Me-LLaMA models. Like all existing LLMs, they are susceptible to generating information with factual errors or biased information. To mitigate this, future studies could incorporate methodologies like reinforcement learning from human feedback (RLHF).[46] This approach could align the models' responses more closely with human values and ensure they are grounded in factual medical knowledge. Another limitation is the current token handling capacity, capped at 4096 tokens, which is a constraint inherited from the backbone LLaMA2 model. Addressing this limitation could involve extending the models' capability to handle longer contexts. This could be achieved by integrating advanced attention techniques, such as sparse local attention,[47] that are able to handle extensive contexts.

**DATA AVAILABILITY**

All datasets employed in the continual pre-training process and evaluation are accessible from their original published venues. The PubMed Central and PubMed Abstracts subset from The Pile are available at https://huggingface.co/datasets/EleutherAI/pile. MIMIC-IV and MIMIC-CXR datasets can be accessed under the PhysioNet Credentialed Health Data Use Agreement 1.5.0 at https://physionet.org/content/mimic-iv-note/2.2/ and https://physionet.org/content/mimic-cxr/2.0.0/ respectively. The RedPajama data is open-released at https://huggingface.co/datasets/togethercomputer/RedPajama-Data-1. Alpaca data is openly released at: https://github.com/tatsu-lab/stanford_alpaca. Dolly data is openly released at: https://huggingface.co/datasets/databricks/databricks-dolly-15k. Share GPT data can be accessed at: https://huggingface.co/datasets/anon8231489123/ShareGPT_Vicuna_unfiltered. The clinical instruction tuning data based on MIMIC-IV and MIMIC-CXR can be accessed under the PhysioNet Credentialed Health Data Use Agreement 1.5.0 through: https://huggingface.co/clinicalnlplab. The Medical Flash Cards and wikidoc QA datasets can be accessed at https://huggingface.co/medalpaca. Other remaining instruction tuning data can be openly accessed at: https://huggingface.co/clinicalnlplab. Me-LLaMA 13B and Me-LLaMA 70B models can be accessed at: https://physionet.org/content/me-llama/1.0.0/, subject to the completion of a credentialed health data use agreement.

**CODE AVAILABILITY**

The code used for evaluation is available at: https://github.com/BIDS-Xu-Lab/ Me-LLaMA.

**ACKNOWLEDGEMENT**

This work received support from the National Institutes of Health (NIH) under grant numbers: 1RF1AG072799, 1R01AG078154, R01AG073435, R01LM013519, RF1AG084178,

R01AG083039, R01CA284646, R01AI172875, R01AG080991, R01AG080624, R01AG080429, 1K99LM01402, 1K99LM014614-01, NIH/NCATS UL1 TR001427, CDC U18 DP006512, and Patient-Centered Outcomes Research Institute (PCORI) under grant numbers: PCORI RI-FLORIDA-01-PS1, PCORI ME-2018C3-14754. We express our sincere appreciation to the creators of datasets such as the MIMIC, the Pile, and RedPajama for making these valuable resources available to the research community. We extend our gratitude to the UF Research Computing team, under the leadership of Dr. Erik Deumens, for their generous provision of computational resources through the UF HiperGator-AI cluster.

## AUTHOR CONTRIBUTION

QX contributed to the conceptualization of the study, conducted the literature search, developed the methodology, contributed to the software development, carried out validation processes, and was primarily responsible for writing the original draft of the manuscript. QC contributed to the conceptualization of the study, developed the methodology, and contributed to reviewing and editing the manuscript. AC played a key role in data curation and project administration, overseeing the planning and execution of research activities, and contributing to reviewing and editing the manuscript. CP, YH, FL, XP, JH, JZ, VK, XZ and LQ were instrumental in software development and validation and reviewing the manuscript. HH took charge of visualization, specifically in the preparation of figures to support the study's findings, involved in the discussion and reviewing the manuscript. LOM and YW were involved in the discussion, review, and editing of the paper. DS was involved in the discussion, human evaluation, and reviewing the manuscript. HX and JB provided overall supervision for the project, including study design, execution, and evaluation, coordination of study team and resources, and thorough review and revision of the manuscript. All authors reviewed the manuscript critically for scientific content, and all authors gave final approval of the manuscript for publication.

## COMPETING INTEREST

The authors have no financial or non-financial conflicts of interest to disclose.

# APPENDIX

## 1 Medical Instruction tuning Data

Table A1 shows the prompt used for each instruction tuning dataset.

**Table A1**. The prompt for each dataset.

| Data | Prompt |
| --- | --- |
| General domain data | "Below is an input that describes a task, maybe paired with a context that provides further information. Write a response that appropriately completes the request. INPUT:{Text} CONTEXT:{Text} OUTPUT:" |
| Medical conversation data | "Given a medical query, provide a concise and clear answer based on the patient's description. INPUT: {text} OUTPUT:" |
| MedInstruct | "Below is an input that describes a medical task, maybe paired with a context that provides further input information. Write a response that appropriately completes the request. INPUT: {text} CONTEXT: {text} OUTPUT:" |
| Medical flash cards | "If you are a medical professional, answer this question truthfully. INPUT: {text} OUTPUT:" |
| MEDIQA | "Answer the input medical question based on the given context. INPUT: {text} CONTEXT: {text} OUTPUT:" |
| MedicationQA | "Answer this medical question truthfully. INPUT: {text} OUTPUT:" |
| LiveQA | "Given a medical query, provide a concise and clear answer based on the given details. INPUT: {text} OUTPUT:" |
| Patient Information QA | "Answer this medical question truthfully. INPUT: {text} OUTPUT:" |
| GuidelineQA | "If you are a medical professional, answer this question truthfully. INPUT: {text} OUTPUT:" |
| Summarization | "Given an abstract of a biomedical paper, generate the title. INPUT: {text} OUTPUT:" |
| Pubmed sentence generation | "The task is to generate the subsequent sentence in a piece of biomedical literature based on the existing content. INPUT: {text} OUTPUT:" |
| Guideline sentence generation | "Write the next part for a clinical guideline. You're given a piece of the guideline, and your task is to continue it. The new part should match the style and detail of the original and be medically correct. INPUT: {text} OUTPUT:" |
| Summarization | "Given an abstract of a biomedical paper, generate the title. INPUT: {text} OUTPUT:" |
| Topic prediction | "The task is to assign MeSH (Medical Subject Headings) terms to a given piece of biomedical literature. The input is the title and abstract of a biomedical literature. The output is a list of applicable MeSH terms. INPUT: {text} OUTPUT:" |
| Causal relation detection | "For the following statement, determine whether it offers: 1) Strong Advice: The statement gives a clear and assertive recommendation or guidance, or 2) Weak Advice: The statement provides a suggestion or mild recommendation but is not very assertive, or 3) No Advice: The statement doesn't offer any recommendation or guidance. INPUT: {text} OUTPUT:" |
| Relation extraction | "Given a medical question, provide the answer to determine the relation between two medical terms in the question. INPUT: {text} OUTPUT:" |
| MIMIC radiology | "The task is to generate the radiology impression based on radiology findings from a patient's clinical note. The input is the radiology findings from a patient's clinical note. Generate an impression accordingly. INPUT: {text} OUTPUT:" |
| MIMIC disease multiple-choice | "The task is to determine whether a patient suffers from certain diseases, and you need to choose the right answer from the choices. The input is the clinical note of a patient. Please determine which of the following disease(s) occurred during the patient's hospital stay, according to the clinical note in the input: A:{text} B:{text} C:{text} D:{text} Output format: The output should be A, B, C, or D only. INPUT: {text} OUTPUT:" |
| MIMIC disease QA | "The task is to determine whether a patient suffers from certain diseases. The input is the clinical note of a patient. Please respond with all of the disease(s) that occurred during the hospital stay, according to the clinical note in the input. INPUT: {text} OUTPUT:" |
| MIMIC disease binary classification | "The task is to determine whether a patient suffers from certain diseases. The input is the clinical note of a patient. Please determine whether all of the following disease(s) occurred during the patient's hospital stay: {text}. Answer with True or False only. INPUT: {text} OUTPUT:" |
| MIMIC mortality | "The task is to determine whether the patient died while in the hospital. The input is the clinical note of a patient. Using the information in the input, determine whether the patient died while in the hospital. INPUT: {text} OUTPUT:" |
| MIMIC manual QA | "The task is answering a question based on a clinical note in the input and your knowledge. The input is the clinical note of a patient. Please answer the question: {text} INPUT: {text} OUTPUT:" |

## 2 Medical Evaluation Benchmark

Table A2 shows the prompt for each dataset used in the NLP evaluation benchmark.

**Table A2.** The prompt for each dataset in our evaluation benchmark.

| Data | Prompt |
|---|---|
| PubMedQA | "Your task is to answer biomedical questions using the given abstract. Only output yes, no, or maybe as answer. INPUT:{Text} CONTEXT:{Text} OUTPUT:" |
| MedQA | "You are a medical doctor taking the US Medical Licensing Examination. You need to demonstrate your understanding of basic and clinical science, medical knowledge, and mechanisms underlying health, disease, patient care, and modes of therapy. Show your ability to apply the knowledge essential for medical practice. For the following multiple-choice question, select one correct answer from A to E. Base your answer on the current and standard practices referenced in medical guidelines. Question:{text} Options: {text} Answer:" |
| MedMCQA | "You are a medical doctor answering realworld medical entrance exam questions. Based on your understanding of basic and clinical science, medical knowledge, and mechanisms underlying health, disease, patient care, and modes of therapy, answer the following multiple-choice question. Select one correct answer from A to D. Base your answer on the current and standard practices referenced in medical guidelines. Question:{text} Options: {text} Answer:" |
| EmrQA | "Given a medical context and an open-ended question related to it, extract the relevant text segment from the context as an answer. Expected output: Only extract and return the text segment from the provided context that directly answers the question. Do not add any new words. Context:{text} Answer:" |
| i2b2 | "In the sentence extracted from clinical narrative notes, identify all the clinically relevant entities, including problems, tests, and treatments. The required answer format is the same sentence with HTML <span> tags to mark up specific entities. Entity Markup Guide: Use <span class=""problem""> to denote a medical problem. Use <span class="" treatment""> to denote a treatment. Use <span class=""test""> to denote a test Input Text: {text} Output Text:" |
| DDI | "The task is to predict relationship between the given head entity labeled as @DRUG1$ and tail entity labeled as @DRUG2$ within a given sentence, this relation which must be in ('mechanism', 'effect', 'advice', 'int', 'none'). mechanism: this type is used to annotate drug-drug interactions that are described by their pharmacokinetic mechanism. effect: this type is used to annotate drug-drug interactions describing an effect or a pharmacodynamic mechanism. advice: this type is used when a recommendation or advice regarding a drug interaction is given. int: this type is used when a drug-drug interactions appears in the text without providing any additional information. none: there has no drug-drug interactions. INPUT: {text}. OUTPUT:" |
| HoC | "The task is to decide the Hallmarks of Cancer (HOC) taxonomy of the article based on its abstract. The input is an abstract text. There are 10 topics you will need to decide whether the article is related to. Topics: sustaining proliferative signaling, evading growth suppressors, resisting cell death, enabling replicative immortality, inducing angiogenesis, activating invasion and metastasis, genomic instability and mutation, tumor promoting inflammation, cellular energetics, avoiding immune destruction. The output should be topics from the above 10 topics, that are related to the input article. Please note one article can be related to multiple topics. Output format: provide a semicolon-separated list of relevant topics. INPUT:{text} OUTPUT:" |
| MTSample | "TASK: The task is to determine the medical specialty or domain that a medical transcription belongs to. The input is a medical transcription. There are 40 medical specialties or domains, and you need to decide which one is the transcription related to. The medical specialties or domains are: 'Surgery', 'Allergy / Immunology', 'Sleep Medicine', 'Pediatrics - Neonatal', 'SOAP / Chart / Progress Notes', 'Bariatrics', 'Pain Management', 'Lab Medicine - Pathology', 'Dermatology', 'Orthopedic', 'Dentistry', 'Psychiatry / Psychology', 'General Medicine', 'Office Notes', 'Letters', 'Neurosurgery', 'Radiology', 'Cosmetic / Plastic Surgery', 'Nephrology', 'Diets and Nutritions', 'Chiropractic', 'Gastroenterology', 'Cardiovascular / Pulmonary', 'Speech - Language', 'Hospice - Palliative Care', 'Autopsy', 'Endocrinology', 'Emergency Room Reports', 'Discharge Summary', 'ENT - Otolaryngology', 'Urology', 'Physical Medicine - Rehab', 'Neurology', 'Podiatry', 'Ophthalmology', 'Rheumatology', 'IME-QME-Work Comp etc.', 'Hematology - Oncology', 'Consult - History and Phy.', 'Obstetrics / Gynecology'. The output should be only one medical specialty or domain from the above 40 specialties or domains, that is most relevant to the medical transcription. Please note that each medical transcript can only be related to one medical specialty or domain. Output format: provide the name of the medical specialty or domain. INPUT:{text} OUTPUT:" |
| PubMedSum | "The task is to summarize an input biomedical literature in six sentences. The input is a biomedical literature. The output is the summary of an input biomedical literature in six sentences. INPUT:{text} OUTPUT:" |
| MIMIC-CXR | "Derive the impression from findings in the radiology report. INPUT:{text} OUTPUT:" |
| BioNLI | "TASK: Please classify the relationship between the given premise and hypothesis into one of the following labels: entailment, contradiction, or neutral. Return only the label. INPUT:{text} OUTPUT:" |
| MedNLI | "TASK: Please classify the relationship between the given premise and hypothesis into one of the following labels: entailment, contradiction, or neutral. Return only the label. INPUT:{text} OUTPUT:" |

## 3 Complex Clinical Case Diagnosis Task

**Data collecting and processing**: We manually downloaded the PDF files for cases published from January 2021 to December 2022. The content from the "Presentation of Case" section was manually extracted to serve as input for the tested LLMs, while the "Final Diagnosis" section was used as the gold standard for each case's final diagnosis. We concatenated the prompt with the content from the "Presentation of Case" section to form the final input for the LLMs. The prompt we used is as follows:

Your task is to provide at least 10 accurate and distinct patient diagnoses based on the input case report. Key points: 1) Diagnoses are confirmed by clinical or anatomic pathology tests, or sometimes by clinical criteria or expert opinion. 2) You will be informed at the end of the case description if diagnostic tests are being ordered to confirm the diagnosis. Ensure that you provide at least 10 most likely diagnoses, listed in order of likelihood, and cover a wide range of unique possibilities. Follow the guidelines for a generation: 1. Each diagnosis should be precise and unique, ensuring a variety of at least 10 possibilities. 2. List one diagnosis per line. 3. Generate at least 10 differential diagnoses related to the input case report. Think step by step.

***Output format***:

Differential diagnosis: 1. 2. 3. 4. 5. 6. 7. 8. 9. 10.

INPUT:{Text from Presentation of Case}

OUTPUT:

**Automatic evaluation**: Since human evaluation is time and cost-prohibitive, we conducted automatic evaluations using GPT-4o, the current most powerful LLM, following previous studies.[40] For each case, we extracted up to 5 individual diagnoses listed in the differential diagnosis lists predicted by the tested models. We used regular expressions to separate the outputs by newlines and remove any numbering before the diagnoses. Then, we used GPT-4o to determine whether each of these diagnoses matched the ground-truth diagnosis using the following prompt:

As an experienced physician, your task is to identify whether the provided predicted diagnosis is in the true differential diagnosis. Please notice same diagnosis might be in different words. Only return "Y" for yes or "N" for no.

"Predict Diagnosis": {diagnosis predicted by tested models}

"True Diagnosis": {gold standard final diagnosis}

**Human evaluation:** We also had Dr. Shung, a clinician included in our author list specializing in internal medicine, perform a manual evaluation. He checked whether the top-1 and top-5 diagnoses predicted by the tested models matched the gold standard final diagnosis. If a match was found, he labeled it as "1"; otherwise, he labeled it as "0." We then calculated the accuracy score based on these labeled results.